# Price Trackers Inspired by Immune Memory

William O. Wilson*, Phil Birkin* and Uwe Aickelin*
wow, pab, uxa@cs.nott.ac.uk

*School of Computer Science, University of Nottingham, UK

**Abstract.** In this paper we outline initial concepts for an immune inspired algorithm to evaluate price time series data. The proposed solution evolves a short term pool of trackers dynamically through a process of proliferation and mutation, with each member attempting to map to trends in price movements. Successful trackers feed into a long term memory pool that can generalise across repeating trend patterns. Tests are performed to examine the algorithm's ability to successfully identify trends in a small data set. The influence of the long term memory pool is then examined. We find the algorithm is able to identify price trends presented successfully and efficiently.

## 1 Introduction

The investigation of time series data for analysis and prediction of future information is a popular and well studied area of research. Historically statistical techniques have been applied to this problem domain, however in recent years the use of evolutionary techniques has seen significant growth in this area. Neural networks [6] [13], genetic programming [7], and genetic algorithms [3] are all examples of methods that have been recently applied to time series evaluation and prediction.

However the use of immune inspired (IS) techniques in this field has remained fairly limited [9]. IS algorithms have been used with success in other fields such as pattern recognition [2], optimisation [5], and data mining [8]. In this paper we propose an IS approach, using trackers to identify trends in time series data, and take advantage of the associative learning properties exhibited by the natural immune system.

The time series proposed for investigation in this paper is that of price movements (Section 2) and the approach used to identify trends in price data is inspired by the immune memory theory of Dr Eric Bell [1]. His theory indicates the existence of two separately identifiable memory populations which are ideally suited to recognise long and short term trends prevalent in time series data. In Section 3 we discuss this immune memory theory and introduce other immune mechanisms which form part of our algorithm. The algorithm itself is then presented in Section 4. The methodology for testing the algorithm, the results and discussions of the results are documented in Sections 5, 6 and 7 respectively, before concluding in Section 8.

## 2   Analysis and representation of price trends

In our approach price data is converted to price movements over time and presented to the system as an antigen. The change in price at time $t_i$ is calculated as the closing price at time $t_i$ less that of $t_{i-1}$. Price movements are then banded to simplify classification. For example a price rise between $0 and $1 is categorised as a $1 price rise and stored as the antigen Ag = [1]. The classification boundary (in this case 1) can be altered as required depending on the level of detail needed in the evaluation. Price movements are then stored in chronological order within a vector representing the antigen. The antigen provides a historical record of price changes over a particular period. The objective of our algorithm is to identify the trends prevalent within that antigen.

A trend 'T' is defined as a sequence of continuous price changes, whose length exceeds one, that are seen to repeat at least once within the antigen. This paper provides a proof of concept that such a trend detection mechanism is possible.

## 3   Development of long and short term memory

The flexible learning approach offered by the immune system is attractive as an inspiration but without an adequate memory mechanism knowledge gained from the learning process would be lost. Memory therefore represents a key factor in the success of the immune system. A difficulty arises in implementing an immune memory mechanism however, because very little is still known about the biological mechanisms underpinning memory development [11]. Theories such as antigen persistence and long lived memory cells [10], idiotypic networks [4], and homeostatic turnover of memory cells [12] have all attempted to explain the development and maintenance of immune memory but all have been contested. The attraction of the immune memory theory proposed by Dr Eric Bell is that it provides a simple, clear and logical explanation of memory cell development. This theory highlights the evolution of two separate memory pools, 'memory primed' and 'memory revertant' [1], see Figure 1.

Antigen presented by dendritic cells in the lymph node causes naive cells to undergo blast transformation and become activated, increasing proliferative capacity, and responsiveness but becoming short lived in the process due to their instability. This rapidly expanding population forms the short lived memory primed pool. The purpose of this growing pool is to drive the affinity maturation process to cope with the huge diversity in the potential antigen repertoire. These cells migrate to the periphery in an attempt to interact with further antigens. If antigen contact is achieved the memory primed cells terminally differentiate into effector cells to counter the antigen, after which point they die.

The high rate of apoptosis of memory primed cells means most will die during circulation of the periphery, however a small minority that fail to achieve secondary antigen exposure do survive and return to the lymph node to reach a memory revertant state. These cells down-regulate cytokine production and apoptotic pathways and revert back to a naive like state. The key difference to naive cells however is that these revertant cells are able to homeostatically turnover, producing clones to sustain knowledge of an antigen experience over the long term. These two distinct memory pools, and the transfer mechanism between them, represent a key difference to other memory theories, and prove the inspiration for memory development in our algorithm.

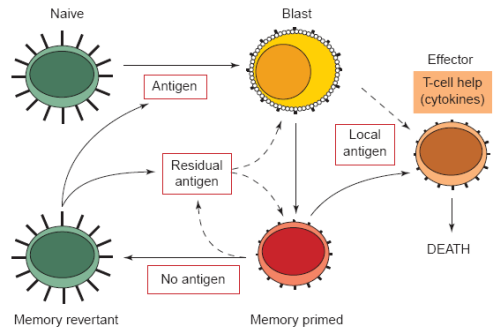

Fig. 1. Immune memory development [1]

In our solution the equivalent of the short term memory pool is generated using a derivative of the popular clonal selection algorithm [5] which proliferates all successfully bound candidates. The short term memory pool evolves through a special form of mutation, and is regulated through apoptosis. Successful candidates from the short term memory pool then transfer to the long term memory pool for permanent storage. This pool can then be utilised during future antigen presentations to aid in identification. These mechanisms are discussed in detail in Section 4.

## 4 An immune inspired trend evaluation and prediction solution

The pseudo code for the proposed Trend Evaluation Algorithm (TEA) is detailed in Program 1. Each of the significant operations in the TEA is then described in the subsequent sections. All parameters noted in these sections have been chosen using educated guesses based on previous experience, no formal sensitivity analysis has been performed to date but will form part of our future work.

### 4.1 Tracker pool construction and initialisation

The TEA comprises a population of individual 'trackers' whose purpose is to identify the price trends located within an antigen. Each tracker is a vector consisting of multiple price change estimates, much like the antigen. The price estimates are generated using a Gaussian distribution and converted to price

**Program 1 . TEA Pseudo Code**

```
Convert oil price data to form antigen 'Ag'
Generate naive tracker pool 'TP'
For generations 1 to N
{
    Present Ag to each tracker 'Ti'
    Calculate affinity 'AF' between Ag and Ti
    Identify optimal match sequence 'MS' in Ti
    Calculate stimulation factor 'SF' of MS
    Calculate match length 'ML' of MS
    If (AF < bind threshold) && (SF or ML > previous SF, ML values)
    {
        Clone Tracker in proportion to ML
        Determine mutation mechanism & mutate clone
        Add clones to TP
    }
    Identify long term (LT) memory candidates from TP
    Transfer successful candidates to LT memory pool
    Apoptose TP
}
```

categories. The initial tracker pool is set at 20 trackers and the length of each tracker is randomly generated on initialisation to contain between one and four price estimates.

### 4.2 Antigen presentation and tracker binding

The algorithm runs for 50 generations. During each generation the latest price change value is calculated and provided to the TEA and added to the current antigen. In generation 'n' the TEA will obtain the $n^{th}$ price change value and present it, along with all previous price values, as an antigen to the current tracker population.

The affinity between the antigen and each tracker is calculating as the numerical difference between the price values in the antigen and the tracker. A bind threshold of zero produces a continuous set, or sub-set, of the tracker that identically matches a part of, or the whole of, the antigen. All possible continuous permutations of the tracker are assessed against the antigen to find the longest matching sequence 'MS' between the two entities. For example, given antigen A1 [0.5, 1, 2] and tracker T1 [1, 2, 1], the MS would be [1,2] after all permutations of T1 and A1 were investigated.

During the binding process the stimulation factor 'SF' for the current MS is determined. This corresponds to the number of times MS is seen to repeat within the antigen. The match length 'ML' of the tracker is calculated as the length of MS. If SF and ML both exceed 1 then the MS represents a recurring trend within the antigen and that tracker is flagged as a candidate for proliferation. To avoid excessive population growth, proliferation candidates only undergo proliferation if their SF or ML values exceed those attained in the previous generation. The tracker is seen to have improved its fitness to an antigen trend (in terms of length of match, or frequency of occurrence) and as such is cloned.

### 4.3 Proliferation and Mutation

All trackers that meet the proliferation criteria are cloned, forming the short term memory pool theorised in Section 3. The number of clones generated during a match is proportional to the ML for that match. This was decided because a proliferation mechanism, using ML as a driver, in conjunction with the mutation mechanism, encourages successful trackers to evolve and lengthen to match ever longer trends.

Clones undergo mutation within the TEA in one of two unique forms, selected randomly with a probability of 0.5.

- Mutation by Extension : Here a new price estimate is generated randomly using a Gaussian distribution and added to the end of the clone.
- Mutation by Shortening : Here a randomly selected price estimate within the tracker is eliminated.

Extension mutation allows the clone, whose parent was a successful match to a trend, to anticipate the next price movement in that trend. The tracker clone evolves to increase the length of it's MS as it tries to detect longer and more complex price trends. During the binding process some trackers will contain redundant price information. Redundant price information is defined as any price values within the tracker that are not included in the MS of that tracker. The shortening mutation permits the trackers a random chance to rid themselves of redundant information and improve the accuracy of the resulting memory pool.

### 4.4 Long term memory transfer

During each generation all trackers undergoing proliferation become candidates for entry into the long term memory pool. Trackers that have a MS not recorded in the memory pool will automatically be transferred into the pool for preservation. Candidates with a MS identical to that of one of the memory trackers will only replace that memory if they contain less redundant information than that memory tracker. The memory pool thus reflects the most efficient matching trackers in the population up to that point in time.

### 4.5 Apoptosis

To ensure the tracker population returns to a stable equilibrium 10% of the current tracker population is selected at random and eliminated. Both high and low affinity trackers have the same probability of death. If the population falls below its minimal limit of 20 the remaining population will automatically clone to repopulate the pool, reflecting homeostatic turnover observed in nature.

To reflect the instability and high death rate prevalent in the short term memory pool clones are eliminated five generations after their creation if they do not improve on their affinity to an antigen trend. This ensures excessive population growth is carefully regulated and a return to a stable population level soon after antigen presentation ceases.

Reviewing the mechanisms within the TEA one can see a close similarity exists to algorithms such as CLONALG [5], however a number of notable differences exist. Compared to CLONALG apoptosis occurs across all population members in the TEA, not just the lowest affinity members. In addition, due to it's specialised nature, mutation in the TEA is not directly related to affinity fit. TEA also proliferates all bound trackers to form the short term memory pool, encouraging diversity in the search space. The process in CLONALG is more elitist, as only the 'n' fittest population members are proliferated and mutated, and from these only the best fitting clone becomes a memory candidate. All remaining clones are eliminated. In essence CLONALG skips the short term memory pool stage as it looks to find the best fitting candidate using the minimum of resources. In comparison the TEA maintains the population of clones in order to match and anticipate patterns arising in the data fed live to the system.

## 5 Testing Methodology

### 5.1 Methodology

In order to test the ability of the TEA to identify trends in a data series, a simple antigen 'A' was constructed. 'A' contains 20 fictitious price movements, and 8 trends, T1 to T8. These represent the complete set of trends in A in accordance with the definition described in Section 2. The antigen and trends T1 to T8 are listed in Table 1.

Table 1. Antigen data sets with observed trends.

| Antigen | Price Movements |
|---|---|
| A | [ 1, 2, 1, -0.5, 1, 2, 1, 0.5, -0.5, 0.5, 2, 1, 2, -0.5, 2, 1, 2, -0.5, 1, 1.5 ] |
| A1 | [ 1, 2, 1, -0.5, 1, 2, 1, 0.5, -0.5, 0.5 ] |
| A2 | [ 2, 1, 2, -0.5, 2, 1, 2, -0.5, 1, 1.5 ] |
| Trends | |
| T1 | [ 1, 2 ] - seen in A, A1 and A2 |
| T2 | [ 1, 2, 1 ] - seen in A and A1 |
| T3 | [ 2, 1 ] - seen in A, A1 and A2 |
| T4 | [ 1, 2, -0.5 ] - seen in A and A2 |
| T5 | [ 2, -0.5 ] - seen in A and A2 |
| T6 | [ 2, 1, 2 ] - seen in A and A2 |
| T7 | [ 2, 1, 2, -0.5 ] - seen in A and A2 |
| T8 | [ -0.5, 1 ] - seen in A |

To assess the ability of the TEA to associate new novel antigen with those experienced during past presentations we split antigen A at the mid point into two subsets A1 and A2, both of length 10. A1 represents the training data set from which the TEA will develop a long term memory of trends associated with A1. A2 represents the testing data set which the TEA will have to examine in the light of information preserved from the experience of A1.

A1 contains three simple trends, T1, T2 and T3. They are closely related, in terms of the price movements they contain, so presenting A1 to the TEA represents a simple challenge to ensure the TEA operates correctly.

A2's purpose is to test the ability of the TEA to handle a more complex antigen with more diverse trends. A2 comprises 6 trends, T1 and T3 as were noted in A1, in addition to four new trends T4 to T7. Compared to A1 we have increased the number of trends from three to six and increased their length and diversity, making it more difficult for the TEA to find all the trends in A2.

It is hypothesized that although trends T4, T5, T6 and T7 are more complex to identify from knowledge of A2 alone, after experiencing trends T1, T2, and T3 from A1's presentation, which are related to T4 to T7, the TEA should develop some form of association between the trends leading to an easier recognition of these new patterns. To test this hypothesis we define the following 4 experiments.

In experiment 1 the training set A1 will be presented to the TEA from generations 1 to 10. The testing set A2 is then presented to the TEA from generations 30 to 40. The TEA is run for 50 generations and the experiment repeated and results averaged across 10 runs. The frequency of detection of trends T1 to T7 is recorded across all runs. To give a base line comparison where there is no memory in the system experiment 1 assumes no knowledge of A1 is carried forward in the TEA during A2's presentation. At the point of A2's presentation the tracker population is replaced by the random tracker population created in generation 0. The TEA therefore has to learn to recognise trends in A2 from scratch.

Experiment 2 investigates the impact of incorporating feedback from the long term memory pool into the TEA. We repeat the previous experiment, but the tracker population at generation 30 is repopulated using clones from the long term memory pool. We identify whether any association properties become apparent in the TEA by examining the frequency with which the trackers in the long term memory pool have detected the trends in A2 as compared to experiment 1.

Experiment 3 investigates the issue of scalability in the TEA. Experiments 1 and 2 present antigen sub sets of only 10 data items at a time. We now scale up the information presented to evaluate the impact on the TEA's performance. Experiment 3 presents the complete antigen A to the TEA from generation 1 to 20, doubling the size of the information presented. Results in terms of population sizes, trend detection rates and memory pool efficiency are then to be compared with experiments 1 and 2.

Experiment 4 compares the performance of the TEA against a random search. Each tracker generated during execution represents a potential search solution;

given the high population levels anticipated in the TEA one could argue that a large randomly generated tracker population would also succeed in identifying the trends prevalent in an antigen. Experiment 4 generates a random population of trackers, whose size is approximately equivalent to the population levels generated during experiments 2 and 3 to examine whether the TEA performs better than a random search in terms of trend detection rates and memory efficiency.

## 5.2 Performance evaluation

The results of the TEA are evaluated as an average across 10 runs. The performance of the algorithm is assessed using two measures i) the number of trends identified against the maximum available for detection and ii) the efficiency of the trackers in the long term memory pool to map to the trends. Efficiency can be measured as the number of price change values included in the memory tracker that are not contained within the match sequence 'MS'. For example if the trend to be found was [2.0, 2.5] and the best fitting tracker was [2.0, 2.5, 3.0] the price value 3.0 within the tracker is redundant given the MS of [2.0, 2.5]. The degree of efficiency, or to be more precise inefficiency, would therefore be calculated as 1 over 3, or 33%. The TEA was written in C++ and run on a windows machine with a Pentium M 1.7 Ghz processor with 1.0 Gb of RAM.

## 6 Results

The results of experiments 1 to 3 are discussed in the following sections and are listed in Table 2 whilst those of experiment 4 are found in Table 3. TEA execution times varied from approximately 40 to 50 seconds for experiments 1 and 2, and 7 to 8 minutes for experiment 3.

Table 2. Detection rate and memory efficiency results.

| Experiment | Trend Detection Frequency | | | | | | | | Total | Detection Rate |
| --- | --- | --- | --- | --- | --- | --- | --- | --- | --- | --- |
| | T1 | T2 | T3 | T4 | T5 | T6 | T7 | T8 | | |
| 1 | 10 | 10 | 10 | 6 | 2 | 1 | 0 | n/a | 39 | 55.7% |
| 2 | 9 | 9 | 10 | 9 | 9 | 7 | 3 | n/a | 56 | 80.0% |
| 3 | 10 | 10 | 10 | 10 | 9 | 10 | 8 | 10 | 77 | 96.3% |

| Experiment | Redundant memory values | | | | | | | | Total | Inefficiency Rate |
| --- | --- | --- | --- | --- | --- | --- | --- | --- | --- | --- |
| | T1 | T2 | T3 | T4 | T5 | T6 | T7 | T8 | | |
| 1 | 0 | 2 | 0 | 0 | 0 | 1 | 0 | n/a | 3 | 3.2% |
| 2 | 0 | 0 | 0 | 0 | 0 | 2 | 1 | n/a | 3 | 2.1% |
| 3 | 0 | 0 | 0 | 0 | 0 | 0 | 1 | 3 | 4 | 2.0% |

## 6.1 Experiment 1. No Long term memory pool interaction

In accordance with Section 5 A1 was presented to the TEA from generations 1 to 10. The tracker population at generation 30 was replaced by the randomly generated tracker population from generation 1. A2 was then presented from generations 30 to 40. Figure 2 illustrates the total tracker population in response to these presentations, whilst Figure 3 shows the population of trackers that specifically match trends T1 to T7.

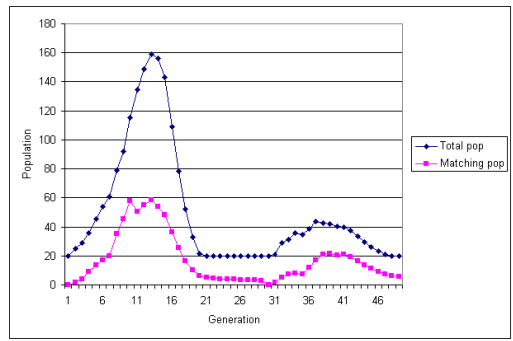

Fig. 2. Total Tracker and total matching tracker populations with no memory feedback

Regarding the presentation of A1, Table 2 shows the TEA is able to identify and develop memory trackers that map with 100% success to trends T1, T2 and T3 for each of the 10 runs. There are no redundant price values in the memory pool resulting in 100% memory efficiency. However the TEA is less successful in indentifying trends T4 to T7 from the subsequent presentation of A2.

The secondary response in Figure 2 is minimal because no memory of the trends from A1 are carried forward in the system, resulting in the TEA having to relearn the trends presented. This led to a poor mapping to A2's trends due to their increased number and complexity.

Trends T1 and T2 were again recognised within A2 and the new trend T4 was identified with 60% success across the 10 runs, however the remaining trends (T5, T6 and T7) were only rarely detected. In total 39 (55.7%) of the 70 possible trends were found across the 10 runs, with 3.2% memory inefficiency.

## 6.2 Experiment 2. Long term memory pool interaction

In this experiment the tracker population is replaced with clones from the memory pool in generation 30. This provides the potential to learn from the trends memorised in response to A1, to create associations with the novel trends in A2. Table 2 shows feedback from the memory pool has a significant impact on the performance of the TEA compared to experiment 1. The total number of trends now mapped by memory trackers increases by 43.6% to 56 trends, giving a detection rate of 80% compared to the previous coverage of 55.7%. The TEA is now able to consistently detect trends T4, T5, and T6 and even manages to identify the elusive T7 with a 30% success rate. Memory inefficiency fell to 2.1% with 3 redundant price values included in the memory population.

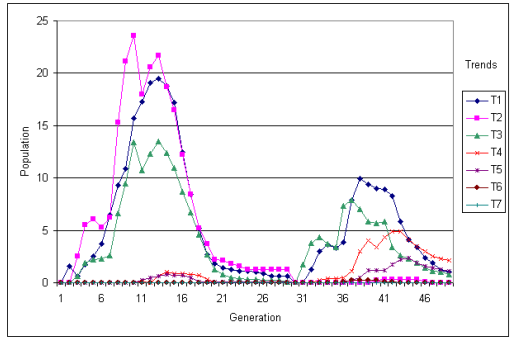

Fig. 3. Trackers matching trends T1 to T7 with no memory feedback

It should be noted however that due to apoptosis during run 4 of the experiment a number of important trackers were eliminated before they had a chance to bind. This resulted in the TEA failing to detect 6 of the 7 available trends in this run. Omitting this unusual occurrence from our analysis would have boosted the current 80% detection rate to 87%.

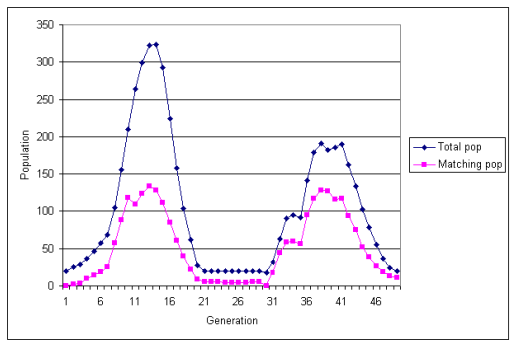

Fig. 4. Total Tracker and total matching tracker populations with memory feedback

Figures 4 and 5 show the total tracker population levels and tracker populations that match the specific trends T1 to T7. Figure 4 shows a more pronounced secondary response to A2 compared to that in Figure 2, with the maximum pop-ulation rising to 191 trackers compared to that of 44 in experiment 1. Looking at the population of trackers that map to specific trends (Figure 5) we see evidence of stronger responses to the trends in A2, as seen in Figure 3. Thus knowledge of the trends seen from A1's presentation have improved the TEA's recognition of new, novel trends that have some association with those previously seen. This leads to the 43.6% improvement in trend detection.

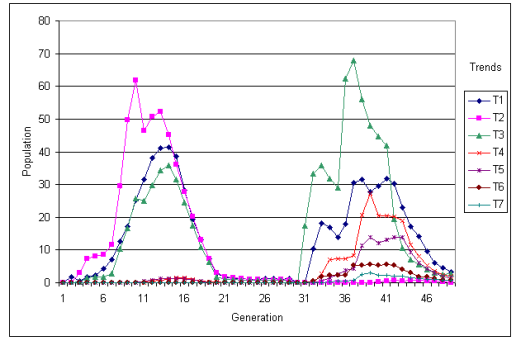

Fig. 5. Trackers matching trends T1 to T7 with memory feedback

This experiment was repeated to examine the impact of removing the shortening mutation function from the TEA. Considering presentation of A1 the tracker population levels reached a slightly higher peak of 218 compared to 191 with no shortening, however the impact on the quality of the memory pool was significant. Whilst the TEA's detection rate for trends T1 and T2 varied insignificantly, without shortening the detection rate for T3 fell from 100% to 10%, T3 was undetected in 9 of the 10 execution runs. Of more concern was the fact that the resulting memory pool contained 33 redundant price values compared to the 100% memory efficiency found through using mutation by shortening. It is clear that the shortening mutation is vital for the proper performance of the TEA.

6.3   Experiment 3. Antigen scalability

To assess scalability antigen A was presented to the TEA from generations 1 to 20. Scanning A we see a new trend T8 becomes apparent when we combine subsets A1 and A2. Detection of this trend would not be possible in any of the previous experiments because its occurrence in A1 and A2 does not satisfy the definition of a trend in those individual sub sets. This highlights an issue with the approach as the point of split in the antigen has an impact on the potential number of trends to be detected in the sub parts of that antigen, this issue is addressed later.

The tracker population reaches a maximum of 2,244 trackers compared to the maximum population in experiment 2 of 323. The memory pool created is able to successfully map to 77 of the 80 possible trends across the 10 runs (96.3% coverage). The TEA failed to find T7 twice and T5 once. Memory inefficiency dropped to 2% as 4 excess price values were noted in the memory pool.

### 6.4 Experiment 4. Comparison with random search

From experiments 2 and 3 approximately 1,000 and 4,000 trackers respectively were generated by the TEA in order to generate the memory pool of solutions. To compare the results of experiments 2 and 3 with a random search, a random population was generated consisting of 1,000, 4,000, 10,000 and 20,000 trackers. Given the longest trend (T7) has four price values, and can be found by the TEA with no data redundancy, each random tracker had a randomly determined length between one and four. The randomly generated population would then be mapped to antigen A, and the memory trackers compared to those of experiments 2 and 3 to see whether the TEA can outperform a purely random search. Results are shown in Table 3, ticks indicate the trend was found, crosses indicate the trend was not detected.

Table 3. Trends detected using a random search.

| Pop Size | Analysis of Trends Detected | | | | | | | | Total |
| --- | --- | --- | --- | --- | --- | --- | --- | --- | --- |
| | T1 | T2 | T3 | T4 | T5 | T6 | T7 | T8 | |
| 1,000 | ✓ | X | ✓ | X | ✓ | X | X | n/a | 3 |
| 4,000 | ✓ | ✓ | ✓ | ✓ | ✓ | X | X | ✓ | 6 |
| 10,000 | ✓ | ✓ | ✓ | ✓ | ✓ | ✓ | X | ✓ | 7 |
| 20,000 | ✓ | ✓ | ✓ | ✓ | ✓ | ✓ | X | ✓ | 7 |

With a randomly generated population of size 1,000 only 3 of the 7 trends T1 to T7 were detected. The random search failed to find trends T2, T4, T6 and T7. In comparison, during experiment 2 the TEA found 6 trends consistently, missing only T7 70% of the time. The detection rate of the TEA is twice that of the random search with just 1,000 trackers.

With 4,000 random trackers 6 of the 8 trends are found, trends T6 and T7 were undetected by the random search. Increasing the random population size to 10,000 trackers, 7 of the 8 trends are detected as T6 is now found. The random search fails to find T7, even if we increase the tracker population to 20,000. This contrasts to experiment 3 where the TEA, with only 4,000 trackers, can generate a memory pool that detects T1, T2, T3, T4, T6 and T8 every time across all 10 runs, and T5 and T7 9 and 8 times out of 10 respectively. The TEA therefore outperforms a random search.

### 7 Discussion

From experiment 1 it is seen that the TEA can evolve a population of trackers that generate a memory pool able to successfully map to trends in a simple data set (such as A1) with 100% accuracy and efficiency. Increasing the number and complexity of the trends to be found, as was achieved through the presentation of A2, causes the algorithm to struggle to identify these potential trends.

Without knowledge of the trends from A1 being carried forward in the system, detection rates of the TEA to the more complex trends falls significantly. This can be corrected in the TEA by increasing the degree of proliferation to raise the detection rate in the system. But what is of interest to us in this paper is whether the TEA can learn, through feedback from its long term memory pool, to associate what it has memorised from previous experiences to aid in the investigation of new novel antigen. Comparing the results of experiments 1 and 2 we see incorporation of the memory pool has a beneficial effect on the ability of the TEA to map to and memorise trends in a more complex antigen. Compared to its naive counterpart the inclusion of the long term memory pool boosts trend recognition from 55.7% to a potential 87.%, whilst inefficiency in the memory pool is kept consistently low at 2.1%.

The reason for this improvement can be seen if we view the trends within the antigen subsets A1 and A2, as shown in Table 1. Trends T1, T2 and T3, located within A1, have price change combinations involving rises of $1 or $2. Recognition and development of memory trackers associated with these trends would assist the TEA in identifying trends T4, T6 and T7 in A2 as they too have price combinations that involve price rises of $1 and $2. If memory trackers can be successfully evolved to map to these trends during presentation of A1, as was shown in experiment 2, then the TEA can utilise that knowledge and associate new novel trends with those already seen, instigating a more successful response. Without the ability to associate new experiences with past knowledge the performance of the TEA declines significantly, as expected.

Although the antigen investigated here is very small and simplistic, it is important for the TEA to scale to handle larger antigens. Experiment 3 gives us an indication of the scalability of the system as antigen sizes increase. Comparing test experiments 2 and 3 we see increasing the antigen size by 100% from 10 to 20 causes the maximum tracker population to increase from 323 trackers to 2,244, leading to an exponetial growth problem. Splitting antigen A into it's two component parts, as done in experiment 3, results in significantly lower population sizes whilst still maintaining a high detection rate. This is only possible if we carry forward the long term memory pool and feed it back into the tracker population to assist in future antigen recognitions. In this way we can avoid the scalability issue whilst maintaining a high degree of accuracy in the TEA.

However, from test 3 it was evident that separating antigen A at the mid point results in trend T8 now not being recognised as a trend within the component parts A1 and A2. T8 exists within A1 and A2 but is not repeatedly stimulated so has a SF of 1, therefore it does not conform to the definition of a trend in either A1 or A2. To avoid this issue the algorithm could be re-run with alternative split points to generate an overall memory pool; this will be investigated in future work. From analysis in experiment 4 we can also conclude that the TEA performs significantly better than a random search in identifying trends prevailing in a small data set.

## 8 Conclusion

This paper presents an immune inspired algorithm that is successful in identifying trends in a small simple data set. The authors theorise that these techniques can be expanded and applied to larger time series data sets to identify trends over time. Potential scalability issues can be addressed by breaking the data into more manageable subsets, so long as memory generated from previous presentations is fed back into the TEA prior to new data presentation. Using this approach the algorithm can learn through association from past experiences to maintain a high success rate in detecting and recording prevalent trends.

## References

1. E. B. Bell, S. M. Sparshott, and C. Bunce. CD4+ T-cell memory, CD45R subsets and the persistence of antigen - a unifying concept. Immunology Today, 19:60–64, February, 1998.
2. J. H. Carter. The immune system as a model for pattern recognition and classification. Journal of American Medical Informatics Association, pages 28–41, January 2000.
3. S. H. Chen. Genetic Algorithms and Genetic Programming in Computational Finance. Kluwer Academic Publishers: Dordrecht, 2002.
4. D. Chowdhury. Immune networks: An example of complex adaptive systems. In Artificial Immune Systems and their Applications, D. Dasgupta (ed), pages 89–104, 1999.
5. L. N. de Castro and F. J. Von Zuben. Learning and optimization using the clonal selection principle. IEEE Transactions on Evolutionary Computation, 6(3):239–251, 2002.
6. M. Ghiassi, H. Saidane, and D. K. Zimbra. A dynamic artificial neural network model for forecasting time series events. International Journal of Forecasting, 21:341–362, 2005.
7. C. Grosan, A. Abraham, S. Y. Han, and V. Ramos. Stock market prediction using multi expression programming, 2005.
8. T. Knight and J. Timmis. AINE: An immunological approach to data mining. In N. Cercone, T. Lin, and X. Wu, editors, IEEE International Conference on Data Mining, pages 297–304, San Jose, CA. USA, 2001.
9. I. Nunn and T. White. The application of antigenic search techniques to time series forecasting. GECCO, pages 353–360, June 2005.
10. A. S. Perelson and G. Weisbuch. Immunology for physicists. Rev. Modern Phys., 69:1219–1267, 1997.
11. W. Wilson and S. Garrett. Modelling immune memory for prediction and computation. In 3rd International Conference in Artificial Immune Systems (ICARIS-2004), pages 386–399, Catania, Sicily, Italy, September 2004.
12. A. Yates and R. Callard. Cell death and the maintenance of immunological memory. Discrete and Continuous Dynamical Systems, 1:43–59, 2001.
13. G. Zhang, D. E. Patuwo, and M. Y. Hu. Forecasting with artificial neural networks: The state of the art. International Journal of Forecasting, 14:35–62, 1998.